\documentclass[runningheads]{llncs}
\usepackage[T1]{fontenc}
\usepackage{graphicx}
\usepackage{amsmath,amssymb,amsfonts}
\usepackage{algorithmic}
\usepackage{graphicx}
\usepackage{textcomp}
\usepackage{xcolor}
\usepackage{subfig}
\usepackage{gensymb}
\usepackage[numbers]{natbib}
\usepackage{caption}
\usepackage{svg}
\usepackage{mwe}
\usepackage{array}
\usepackage{amsfonts}
\usepackage{amssymb}
\usepackage{verbatim}
\usepackage{amsbsy}
\usepackage{siunitx}
\usepackage{tabularx, booktabs}
\def\BibTeX{{\rm B\kern-.05em{\sc i\kern-.025em b}\kern-.08em
    T\kern-.1667em\lower.7ex\hbox{E}\kern-.125emX}}

\usepackage{soul}
\usepackage{tikz}
\usetikzlibrary{calc}
\begin{document}
\title{UruBots Autonomous Car Team Two: Team Description Paper for FIRA 2024}
%
%
\author{William Moraes\inst{1} \and Juan Deniz\inst{1} \and Pablo Moraes\inst{1} \and Christopher Peters\inst{1} \and Vincent Sandin\inst{1} \and Gabriel da Silva\inst{1} \and Franco Nunez\inst{1} \and Maximo Retamar\inst{1}  \and Victoria Saravia\inst{1} \and Hiago Sodre\inst{1}\and Sebastian Barcelona\inst{1} \and Anthony Scirgalea\inst{1} \and  Bruna Guterres\inst{1} \and André Kelbouscas\inst{1} \and Ricardo Grando\inst{1} }
\authorrunning{UruBots AC Two et al.}
\titlerunning{UruBots AC Two}
%
\institute{Technological University of Uruguay - UTEC}

\maketitle              
\begin{abstract}
This paper proposes a mini autonomous car to be used by the team UruBots for the 2024 FIRA Autonomous Cars Race Challenge. The vehicle is proposed focusing on a low cost and light weight setup. Powered by a Raspberry PI4 and with a total weight of 1.15 Kilograms, we show that our vehicle manages to race a track of approximately 13 meters in 11 seconds at the best evaluation that was carried out, with an average speed of 1.2m/s in average. That performance was achieved after training a convolutional neural network with 1500 samples for a total amount of 60 epochs. Overall, we believe that our vehicle are suited to perform at the FIRA Autonomous Cars Race Challenge 2024, helping the development of the field of study and the category in the competition.

\end{abstract}

\section{Introduction}

The development of autonomous vehicle technologies represents possible improvements in transportation, offering promising solutions for safer and more efficient travel. The 2024 FIRA Autonomous Cars Race Challenge is an opportunity that promotes the development of this area of study by providing a space for showing new autonomous navigation and control approaches. In this context, our team introduces its designed autonomous car, focused on a low cost and light weight issues to attend and compete at the event.

The vehicle, built to RC-car dimensions, incorporates hardware and software development to meet the diverse challenges posed by the race. Central to our design methodology is the implementation of the DonkeyCar software platform. This framework is made for training and deploying self-driving models, which are based on convolutional neural networks in a supervised learning fashion. 

Our aim here is to showcase the caracteristics of our autonomous vehicle and highlight its readiness for the 2024 FIRA Autonomous Cars Race Challenge, demonstrating our team's vision on this kind of vehicle and category. We first introduce our Hardware and Software Setup and then show how our evaluation was carried out at the end.

\section{Hardware}


The autonomous vehicle relies on a selection of certain components to operate accordingly. These components work together to enable autonomous navigation and real-time decision-making. Based on RC (Remote Control) car projects, the vehicle integrates processing units, motor control systems, and vision components to perceive its environment and make informed decisions. Table 1 shows the components used in our vehicle.

\begin{table}[!h]
    \centering
    \begin{tabular}{|c|c|p{8cm}|} 
        \hline 
        \textbf{}& \textbf{Component} & \textbf{Description} \\ \hline 
         & Raspberry Pi 4 model B & This microcontroller was chosen due to its combination of capabilities and cost. Its role in the system includes processing the camera and on-board software. \\ \hline 
         & Shield PCA 9685 & It is a 16-channel, 12-bit PWM driver that communicates via I2C. It allows precise control of the vehicle's actuators, sending signals to the servo motor and the L298N shield, and facilitating communication with the on-board software. \\ \hline 
         & Step Down XL4016 & Used to reduce the LiPo battery voltage from 11V to 5V, suitable for the Raspberry Pi and the PCA 9685 shield. \\ \hline 
         & Servo motor & Used to control the direction of the vehicle. It receives PWM signals from the PCA9685 shield and adjusts the steering angle of the wheels. \\ \hline 
         & DC motors & These motors allow the car to move. Their speed and direction are controlled by PWM signals managed by the L298N motor driver. \\ \hline 
         & Shield L298N & The L298N is a dual H-bridge motor driver that allows independent control of two DC motors. It handles high currents and allows speed and direction control through PWM inputs from the PCA9685. \\ \hline 
         & Camera Logitech c270 & The Logitech c270 is used for real-time image processing and computer vision tasks, allowing the Raspberry Pi to interpret the environment and navigate the car autonomously. \\ \hline 
         & LiPo battery 11.1V(3S) 2200mAh 30C & A Lithium Polymer (LiPo) battery with a capacity of 11.1V, 2200mAh and discharge rate of 30C, that provides lightweight, high-capacity power for the RC car. It powers the motors, Raspberry Pi, and other electronic components. \\ \hline
    \end{tabular}
    \caption{Components of the project}
    \label{tab:Components}
\end{table} 

The chasis of the autonomous car was 3D printed using PLA. The model was based on a project of Thingiverse \cite{thingiverse}. Then it was modified to use DC motors for the rear wheel drive, and some other modifications to use the components mentioned at the Table \ref{tab:Components}. Our proposed vehicle can be seen in the Figure 1.

\begin{figure}
    \centering
    \includegraphics[width=0.6\linewidth]{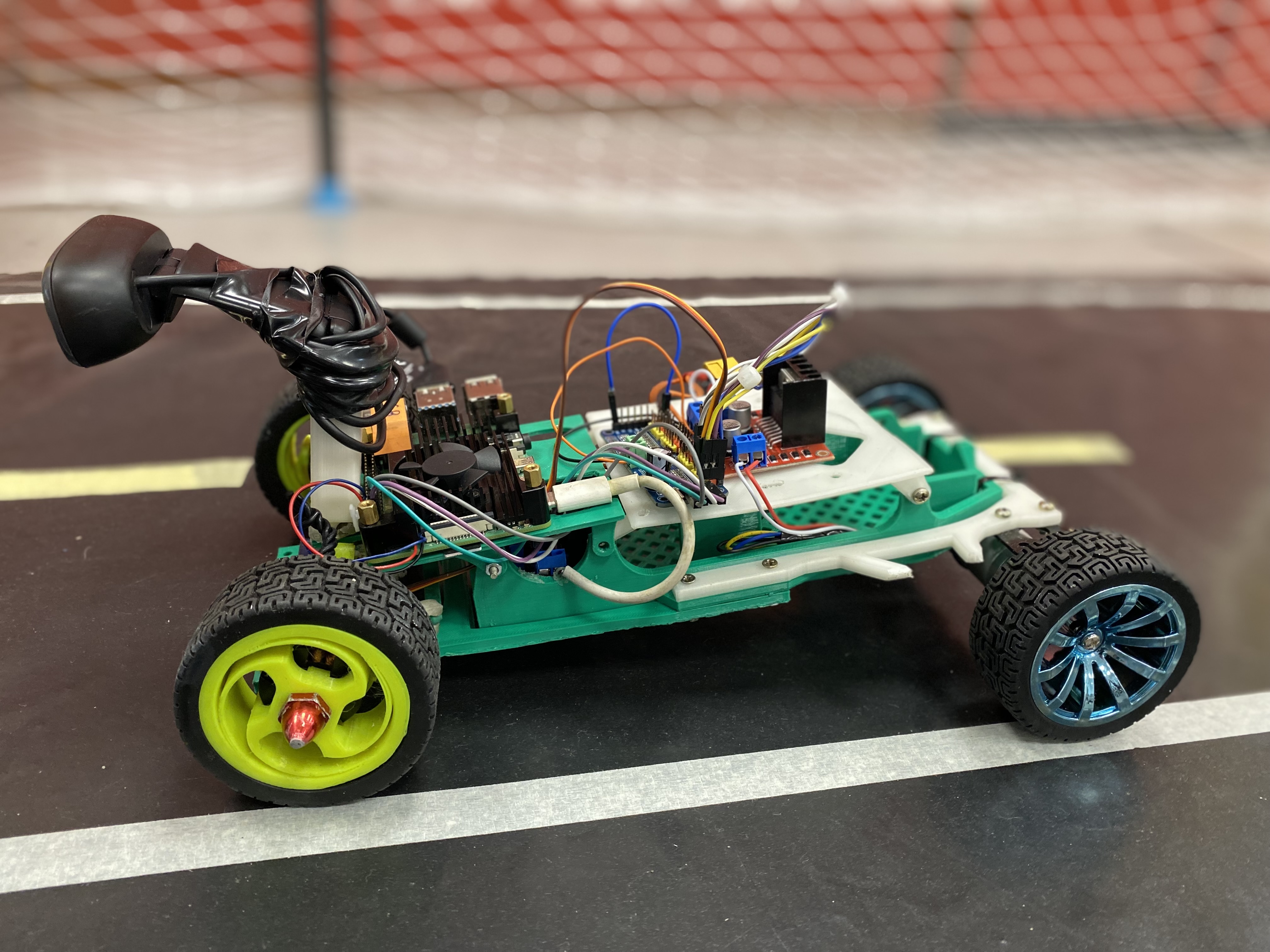}\captionsetup{justification=centering}\label{fig:car}
    \caption{\parbox{0.8\linewidth}{Autonomous car assembled with all components.}}
\end{figure}

\section{Software}


The framework used in this project is based on Donkeycar. Donkeycar is an open-source project designed to facilitate the autonomous vehicles building. Donkeycar gives a solid structure and a set of tools that make development easy to build and program autonomous cars. It can be installed on various single-board computers, in our case was used on a Raspberry Pi 4. The setup includes creating a Donkeycar application from a template, which can be customized to suit specific requirements. The core of the training process involves supervised learning from a collected dataset of a track, where the car learns to drive autonomously based on the user's driving behavior. 


This software is designed for building and training self-driving RC cars, similar to this project. However, this car incorporates some different characteristics from the typical setups used by Donkeycar collaborators, such as a Raspberry Pi 4 and DC Motors. For this project, some adjustments were made to configuration files. Parameters such as steering angle, throttle, communication between components, and certain driving or training functions were configured.

\section{Training}


The data collection involves driving the vehicle while recording camera shots. During this process, the angles of the servomotor and the throttle of DC motors are logged. Subsequently, all recorded actions are saved along with their corresponding labels in a ".json" file.
\begin{figure}
    \centering
    \includegraphics[width=0.6\linewidth]{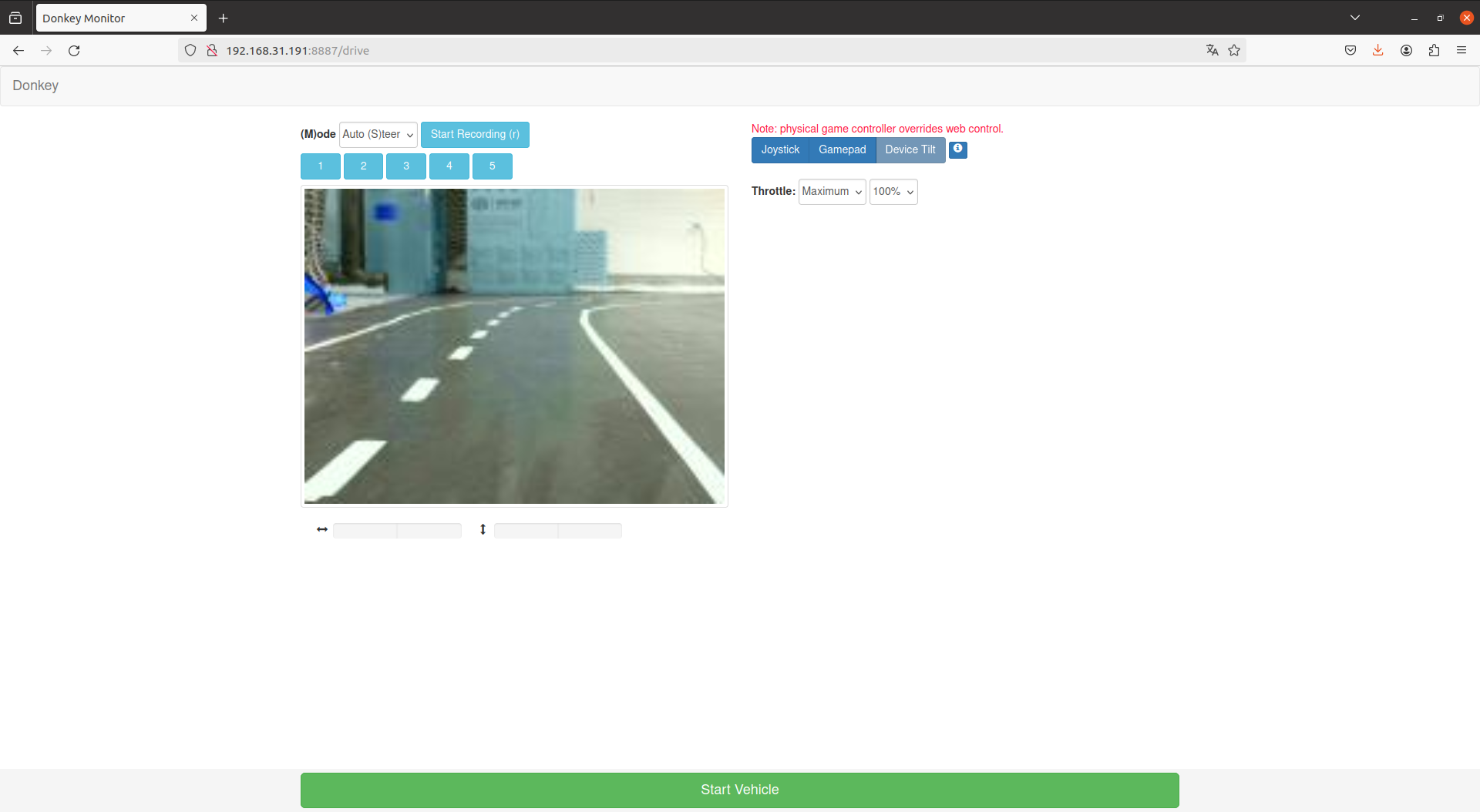}
    \caption{Window print of Donkey Monitor}
    \label{fig:donkeymonitor}
\end{figure}


During the training stage, the collected data is first preprocessed, which includes adjusting the image size and normalization. Subsequently, the model is initialized using a convolutional neural network. The training process begins with the preset 60 epochs, during which the convolutional neural network learns to recognize patterns within the images relevant to the speed and angle of turn parameters. As training progresses, the network gradually adjusts the output parameters.

This model is based on Keras Linear model, that uses these layers:

\begin{itemize}
    \item \textbf{Convolutional Layers (Conv2D):} There are a total of five convolutional layers. These layers are responsible for extracting features from the input images by applying convolutional filters.
    \item \textbf{Dropout Layers:} Six dropout layers are used in total. These layers help prevent overfitting by randomly deactivating a percentage of neurons during training. 
    \item \textbf{Input Layer (InputLayer):} There is one input layer that receives input data in the form of images. 
    \item \textbf{Densely Connected Layers (Dense):} Two densely connected layers are used in total. These layers are fully connected and are used to combine the features extracted by the convolutional layers before the final output. 
    \item \textbf{Flattening Layer (Flatten): }A flattening layer is utilized to transform the output data from the convolutional layers into a suitable format for input into the densely connected layers. 
    \item \textbf{Output Layers (Dense):}  There are two output layers in total. These layers provide the final predictions of the model for the speed and steering angle parameters. 
\end{itemize}


During the execution phase, the model is loaded and prepared to make decisions in real time. To run the trained model module, the 'Donkey Monitor' interface is used, as shown in the Figure \ref{fig:donkeymonitor}. Images from the vehicle camera feed are processed by passing through the model, which then predicts the appropriate driving action—whether to turn or continue straight. This forms the basis for the vehicle's control action, adjusting both speed and steering to adapt to its surroundings. This process operates in a continuous loop, enabling the vehicle to autonomously navigate through its environment while in motion. 

\begin{figure}
    \centering
    \includegraphics[width=0.5\linewidth]{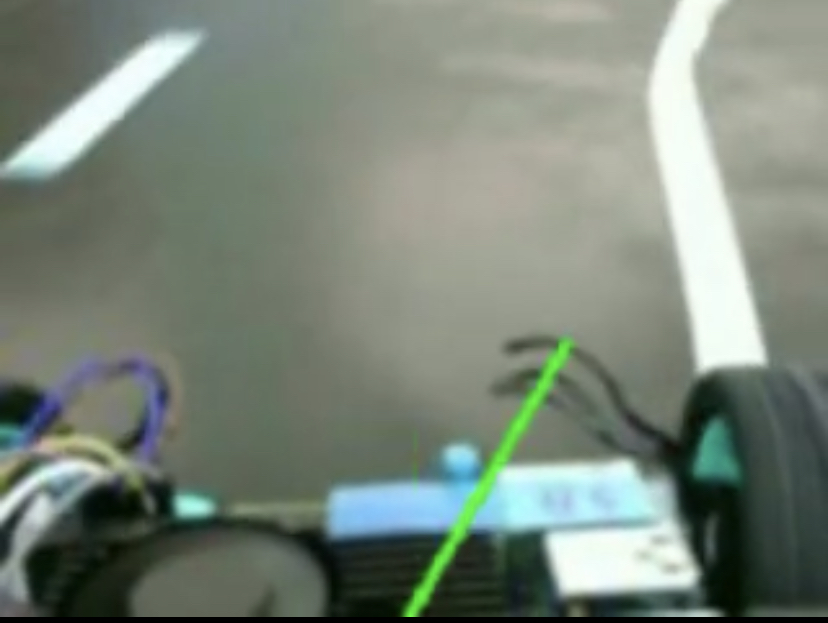}
    \caption{View of car camera showing the next movement vector.}
    \label{fig:vectorcaptura}
\end{figure}

\section{Results}

Our autonomous car adheres to the FIRA Challenge's specifications, ensuring its eligibility for participation. The car's dimensions are within the allowed range, measuring 300mm in length, 200mm in width, and 300mm in height. It is powered electrically and equipped with an Ackerman steering system, which is essential for accurate and responsive maneuvering. Additionally, we selected a Raspberry Pi 4 microcomputer for its processing power, which is crucial for decision-making and navigation in real time and embedded in the vehicle, not requiring processing external to the described hardware.

Our vehicle is built for making it well-suited for the dual challenges of the FIRA competition. The integration of the Raspberry Pi 4 and our modified version of the DonkeyCar software platform allows for robust image processing and autonomous navigation. Furthermore, our use of DC motors and the L298N motor driver ensures reliable speed and direction control. With these features, our autonomous car is fully equipped to handle the race and urban driving tasks, demonstrating its capabilities and our team's expertise in the field of autonomous vehicle technology.

\section{Conclusion}

In this description paper, we show our proposed vehicle to compete at 2024 FIRA Autonomous Cars Race Challenge. Our vehicle was project in such a way that it was cheap to built and  fast at the same time, given its light weight characteristics. We showed that our vehicle manages to complete a proposed race track, surpassing 1 m/s in average. Overall, we concluded that our vehicle has potential to compete at the competition and we look forward for beign part of it and helping the its development.


\begin{thebibliography}{8}
\bibitem{thingiverse}
Midbrink. (2019). RC Car by Oliver Midbrink. Retrieved from \url{https://www.thingiverse.com/thing:4039578} [Accessed: May 22, 2024].
\bibitem{donkeycar}
Donkey Car. (n.d.). Donkey Car Documentation. Retrieved from \url{https://docs.donkeycar.com/}
\end{thebibliography}
\end{document}